\documentclass[10pt,twocolumn,letterpaper]{article}

\usepackage{cvpr}
\usepackage{times}
\usepackage{epsfig}
\usepackage{graphicx}
\usepackage{amsmath}
\usepackage{amssymb}
\usepackage{url}
\usepackage{fixltx2e}
\usepackage{hyperref}
\usepackage{algorithm2e}
\usepackage{enumitem}

\cvprfinalcopy 

\def\cvprPaperID{184} 

\ifcvprfinal\fi
\begin{document}


\title{Stacked Quantizers for Compositional Vector Compression}

\author{Julieta Martinez
\and
Holger H. Hoos\\
University of British Columbia\\
{\tt\small \{julm,hoos,little\}@cs.ubc.ca}
\and
James J. Little
}

\maketitle

\begin{abstract}



Recently, Babenko and Lempitsky~\cite{aq} introduced Additive Quantization (AQ), a generalization of Product Quantization (PQ) where a non-independent set of codebooks is used to compress vectors into small binary codes. Unfortunately, under this scheme encoding cannot be done independently in each codebook, and optimal encoding is an NP-hard problem. In this paper, we observe that PQ and AQ are both compositional quantizers that lie on the extremes of the codebook dependence-independence assumption, and explore an intermediate approach that exploits a hierarchical structure in the codebooks.
This results in a method that achieves quantization error on par with or lower than AQ, while being several orders of magnitude faster.
We perform a complexity analysis of PQ, AQ and our method, and evaluate our approach on standard benchmarks of SIFT and GIST descriptors, as well as on new datasets of features obtained from state-of-the-art convolutional neural networks.
\end{abstract}

\section{Introduction}
Vector quantization has established itself as a default approach to scale applications such as visual recognition and image retrieval. Quantization is usually performed on large datasets of local descriptors (\eg, SIFT~\cite{sift}), or global representations (\eg, VLAD~\cite{vlad} or Fisher vectors~\cite{fisher}). Recent work has also explored the performance-vs.-compression trade-off in state-of-the-art features obtained from deep convolutional neural networks~\cite{gpuretrieval}. Outside the computer vision community, vector quantization is also studied in information theory, multimedia retrieval and unsupervised learning.

Vector quantization is usually posed as the search for a set of \textit{codewords} (\ie, a \textit{codebook}) that minimize quantization error. The problem can be solved in a straightforward manner with the k-means algorithm which, unfortunately, scales poorly for large codebooks. While larger codebooks achieve lower quantization error, the downside is that encoding and search times scale linearly with codebook size.

\begin{figure}
\includegraphics[width=0.5\textwidth, trim=40 80 0 0, clip=true]{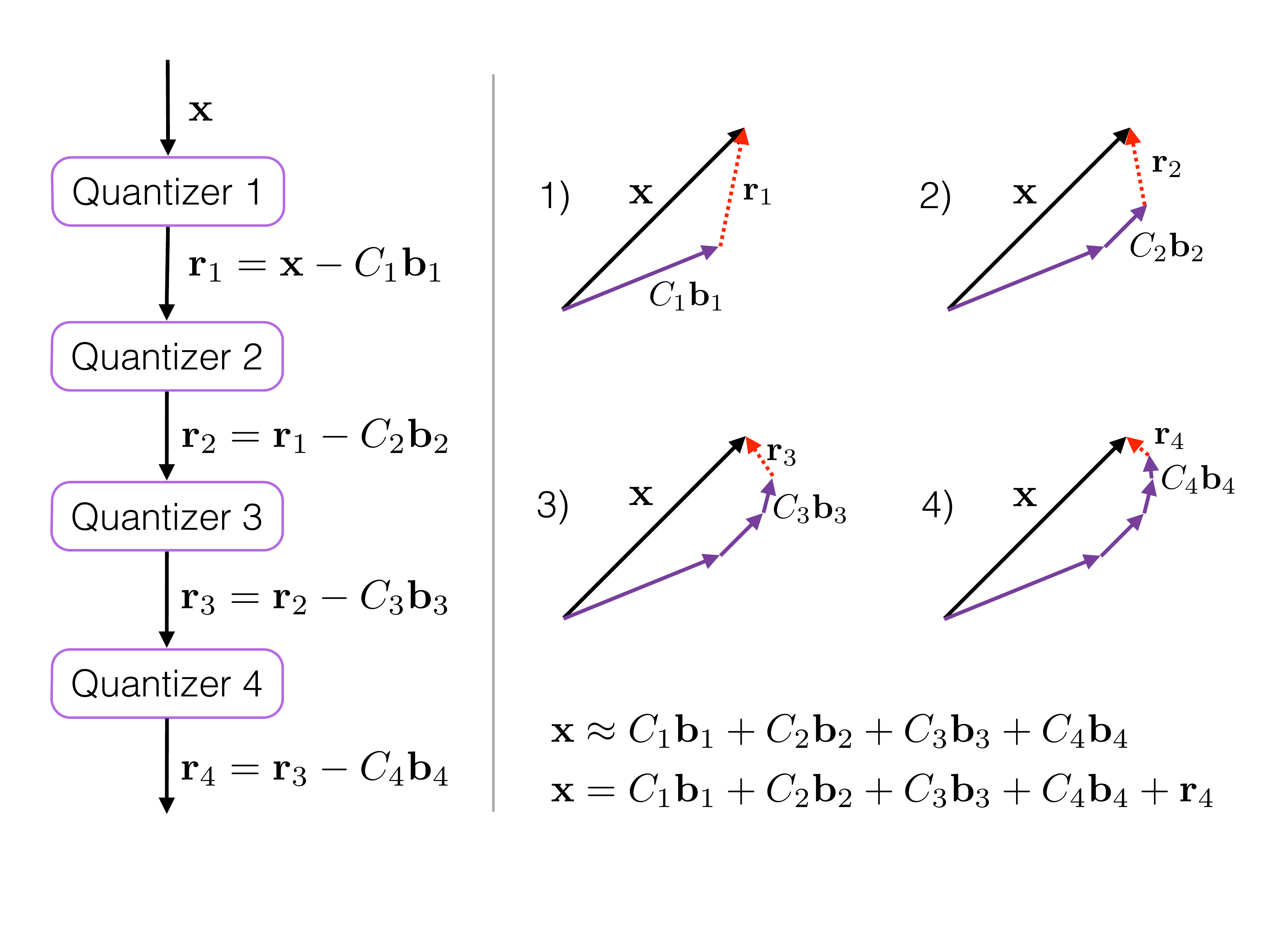}\centering
\caption{Encoding as performed with Stacked Quantizers, shown for 4 subcodebooks. Left: the vector is passed through a series of quantizers, with residuals further encoded down the line. Right: A geometric interpretation of our approach. After recursively encoding residuals, the representation is additive in the encodings, and the quantization error is the remaining residual.}
\label{fig:overview}
\end{figure}

Several algorithms, such as kd-trees and hierarchical k-means, alleviate the search and encoding problems by indexing the codebook with complex data structures~\cite{flann}, achieving sublinear search time as a trade-off for recall. These approaches, however, have a large memory footprint, since all the uncompressed vectors must be kept in memory.

Another line of research considers approaches with an emphasis on low memory usage, compressing vectors into small binary codes.
While for a long time hashing approaches were the dominant trend~\cite{itq, spectral-hashing}, they were shown to be largely outperformed by Product Quantization (PQ)~\cite{pq}. PQ is a compositional vector compression algorithm that decomposes the data into orthogonal subspaces and quantizes each subspace independently. As a result, vectors can be encoded independently in each subspace, and distances between uncompressed queries and the database can be efficiently computed through a series of table lookups.
This combination of small memory footprint, low quantization error and fast search makes PQ a very attractive approach for scaling computer vision applications.

Recently, Babenko and Lempitsky~\cite{aq} introduced Additive Quantization (AQ), a generalization of PQ that retains its compositional nature, but is able to handle subcodebooks of the same dimensionality as the input vectors. With a few caveats, AQ can also be used for fast approximate nearest neighbour search and consistently achieves lower quantization error than PQ. However, since the codebooks are no longer pairwise orthogonal (\ie, no longer independent), encoding cannot be done independently in each subspace.
In~\cite{aq}, beam search was proposed as a solution to this problem, but this results in very slow encoding, which greatly limits the scalability of the proposed solution. 

In this paper, we first analyze PQ and AQ as compositional quantizers, under a framework that makes the simplifying assumptions of PQ w.r.t. AQ rather evident. We next investigate the computational complexity implications resulting from the differences between AQ and PQ, and finally derive an intermediate approach that retains the expressive power of AQ, while being only slightly slower than PQ.

Our approach compares favourably to AQ in 3 ways: (i) it consistently achieves similar or lower quantization error (and therefore, lower error than PQ), (ii) it is \textit{several orders of magnitude faster} and (iii), it is also simpler to implement.


\section{Background and related work}
\label{sect:related}

We introduce some notation mostly following~\cite{ckmeans}. We review the vector quantization problem, the scalability approaches proposed by PQ and AQ, and discuss their advantages and disadvantages.

\subsection{Vector quantization}



Given a set of vectors $\mathcal{X} = \{ \mathbf{x}_1, \mathbf{x}_2, \dots, \mathbf{x}_n \}$, the objective of vector quantization is to minimize the quantization error, \ie, to determine

\begin{align}
\label{eq:quant1}
    \min_{C, \mathbf{b}} \frac{1}{n} \sum_{\mathbf{x} \in \mathcal{X}} \lVert \mathbf{x} - C\mathbf{b} \rVert_2^2,
\end{align}

\noindent where $C \in \mathbb{R}^{d \times k}$ contains $k$ cluster centers, and $\mathbf{b} \in \{0,1\}^k$ is subject to the constraints $\lVert \mathbf{b} \rVert_0 = 1$ and $\lVert \mathbf{b} \rVert_1 = 1$. That is, $\mathbf{b}$ may only index into one entry of $C$. $C$ is usually referred to as a {\em codebook}, and $\mathbf{b}$ is called a {\em code}.

If we let $X = [\mathbf{x}_1, \mathbf{x}_2, \dots, \mathbf{x}_n] \in \mathbb{R}^{d \times n}$ contain all the $\mathbf{x} \in \mathcal{X}$, and similarly let $B = [\mathbf{b}_1, \mathbf{b}_2, \dots, \mathbf{b}_n] \in \{0,1\}^{k \times n}$ contain all the codes, the problem can be expressed more succinctly as determining

\begin{align}
\label{eq:quant2}
    \min_{C, B} \frac{1}{n} \lVert X - CB \rVert_2^2.
\end{align}

Without further constraints, one may solve expression~\ref{eq:quant2} using the k-means algorithm, which alternatively solves for $B$ (typically exhaustively computing the distance to the $k$ clusters in $C$ for each point in $X$) and $C$ (finding the mean of each cluster) until convergence. The performance of k-means is better as the size of the codebook, $k$, grows larger but, unfortunately, the algorithm is infeasible for large codebook sizes (for example, $k=2^{64}$ clusters would far exceed the memory capacity of current machines). The challenge is thus to handle large codebooks that achieve low quantization error while having low memory overhead.


\subsection{Compositional quantization models}

One way of scaling the codebook size looks at compositional models, where smaller subcodebooks can be combined in different ways to potentially represent an exponential number of clusters. Compositional quantization can be formulated similarly to k-means, but restricted to a series of constraints that introduce interesting computational trade-offs. The objective function of compositional quantization can be expressed as

\begin{align}
\label{eq:compquant1}
    \min_{C_i, \mathbf{b}_i} \frac{1}{n} \sum_{\mathbf{x} \in \mathcal{X}} \lVert \mathbf{x} - \sum_i^{m}C_i\mathbf{b}_i \rVert_2^2,
\end{align}

\noindent that is, the vector $\mathbf{x}$ can be approximated not only by a single codeword indexed by its code $\mathbf{b}$, but by the {\em addition} of its encodings in a series of codebooks.
We refer to the $C_i$ as {\em subcodebooks}, and similarly call the $\mathbf{b}_i$ {\em subcodes}.
We let each subcodebook contain $h$ cluster centres: $C_i \in \mathbb{R}^{d \times h}$, and each subcode $\mathbf{b}_i$ remains limited to having only one non-zero entry: $\lVert \mathbf{b}_i \rVert_0 = 1$, $\lVert \mathbf{b}_i \rVert_1 = 1$.
Since each $ \mathbf{b}_i $ may take a value in the range$[1,2, \dots, h]$, and there are $m$ subcodes, the resulting number of possible cluster combinations is equal to $h^m$, \ie, superlinear in $m$. Now we can more succinctly write expression~\ref{eq:compquant1} as

\begin{align}
\label{eq:compquant}
    \lVert X - CB \rVert_2^2 = \lVert X - \left[ C_1, C_2, \dots, C_m \right]  \left[ \begin{matrix} B_1 \\ B_2 \\ \vdots \\ B_m \end{matrix} \right] \rVert_2^2,
\end{align}

\noindent where $B_i = [\mathbf{b}_{i1}, \mathbf{b}_{i2}, \dots, \mathbf{b}_{in}] \in \{0,1\}^{h \times n}$. As we will show next, AQ, PQ and Optimized Product Quantization (OPQ)~\cite{ckmeans, opq} belong to this family of models.

\subsubsection{Product Quantization}
PQ can be formulated right away with Eq.~\ref{eq:compquant} under the constraint that all the subcodebooks be pairwise orthogonal~\cite{ckmeans}:

\begin{align}
	\forall i, j : i \neq j \rightarrow {C}_i^\top C_j = 0_{h \times h},
\end{align}

\noindent that is, $C$ is blockwise diagonal~\cite{ckmeans}:

\begin{align}
	C = \left[ C_1, C_2, \dots, C_m \right] = \left[\begin{matrix} D_1 & 0 & \dots & 0 \\ 0 & D_2 & & 0\\ \vdots & & \ddots & \vdots\\ 0 & 0 & \dots & D_m\end{matrix}\right],
\end{align}

\noindent where the entries $D_i \in \mathbb{R}^{(d/m) \times h}$ are the only non-zero components of $C$. This constraint assumes that the data in $X$ was generated from a series of mutually independent subspaces (those spanned by the subcodebooks $C_i$), which is rarely the case in practice. There are, however, some advantages to this formulation.




\textbf{The subcodebook independence of PQ}
offers 3 main advantages,
\begin{enumerate}[leftmargin=*]
\item Under the orthogonality constraint we can efficiently learn the subcodebooks $\mathcal{C}_i$ by independently running k-means on $d/m$ dimensions. The complexity of k-means is $\mathcal{O}(nkdi)$ for $n$ datapoints, $k$ cluster centres, $d$ dimensions and $i$ iterations. PQ solves $m$ $d/m$-dimensional k-means problems with $h$ cluster centres each, resulting in a complexity of $\mathcal{O}(mnh(d/m)i) = \mathcal{O}(nhdi)$; \ie, training PQ is as complex as solving a k-means problem with $h$ cluster centres.

\item Once training is done, the encoding of the database can also be performed efficiently in $\mathcal{O}(nhd)$ (in line with k-means), which is essential for very large databases.

\item Distance computation between a query $\mathbf{q}$ and a encoded vector $\sum_{i=1}^{m}C_i\mathbf{b}_i$ is efficient because the subcodebooks are orthogonal, and therefore the total distance is equal to the sum of the distances in each subspace~\cite{pq}: $\lVert \mathbf{q} - \sum_{i=1}^{m}C_i\mathbf{b}_i \rVert^2_2 = \sum_{i=1}^m \lVert \mathbf{q}_i - D_i\mathbf{b}_i \rVert^2_2$, where $\mathbf{q} = [\mathbf{q}_1, \mathbf{q}_2, \dots, \mathbf{q}_m]$, and $\mathbf{q}_i \in \mathbb{R}^{d/m}$. These distances can be precomputed for each query and quickly evaluated with $m$ table lookups. This is called Asymmetric Distance Computation in~\cite{pq} and is the mechanism that makes PQ attractive for fast approximate nearest neighbour search.
\end{enumerate}



\subsubsection{Optimized Product Quantization}

One of the main disadvantages of PQ is that $X$ is forced to fit in a model that assumes that the data was generated from statistically independent subspaces. Lower quantization error can be achieved if more degrees of freedom are added to the model. In particular, since rotation is a distance-preserving operation, it seems natural to experiment with codebook rotations that minimize quantization error. In OPQ, the objective function becomes~\cite{ckmeans}

\begin{align}
\label{eq:quantopq}
    \min_{R, C, B} \frac{1}{n} \lVert X - R C B \rVert_2^2,
\end{align}

\noindent where $C$ and $B$ are expanded as in Eq.~\ref{eq:compquant}, and $R$ belongs to the Special Orthogonal Group $SO(d)$. In this sense, PQ is a special case of OPQ where $R$ is the $d$-dimensional identity matrix: $R = I_{d}$. Independently, Ge \etal~\cite{opq} and Norouzi \& Fleet~\cite{ckmeans} proposed an iterative method similar to Iterative Quantization~\cite{itq} that optimizes $R$ in expression~\ref{eq:quantopq}. Notice, however, that the orthogonality constraint is maintained from PQ to OPQ.

Lower quantization error can be achieved if the independence assumption is not enforced, at the cost of more complex encoding and distance computation. These trade-offs were first introduced in~\cite{aq} and called Additive Quantization (AQ). We briefly review AQ here.

\subsubsection{Additive Quantization}

In AQ, the subspaces spanned by the subcodebooks $C_i$ are not mutually orthogonal (\ie, not mutually independent). Formally, and although not explicitly stated in~\cite{aq}, AQ solves the formulation of Eq.~\ref{eq:compquant1} without any further constraints. This makes of AQ a strictly more general model than PQ/OPQ. However, this complexity comes at a cost.

\textbf{The subcodebook dependence of AQ} comes with 3 main disadvantages with respect to PQ/OPQ,

\begin{enumerate}[leftmargin=*]

\item  The distance between a query $\mathbf{q}$ and a encoded vector $\sum_{i=1}^{m}C_i\mathbf{b}_i$ cannot be computed with $m$ table lookups. However, it can be found using the identity

\begin{align}
	&\lVert \mathbf{q} - \sum_{i=1}^{m}C_i\mathbf{b}_i \rVert^2_2 = \notag \\
	&\lVert \mathbf{q} \rVert^2_2 - \sum_{i=1}^m 2 \langle \mathbf{q} , C_i\mathbf{b}_i \rangle + \lVert \sum_{i=1}^m C_i\mathbf{b}_i \rVert^2_2
	\label{eq:dist_aq}
\end{align}

\noindent where the first term is a constant and does not affect the query ranking; the second term can be precomputed and stored for fast evaluation with $m$ table lookups, and the third term can either be precomputed and quantized for each vector in the database (at an additional memory cost), or can be computed on the fly as

\begin{align}
\label{eq:dnorm}
\lVert \sum_{i=1}^m C_i\mathbf{b}_i \rVert^2_2 = \sum_{i}^m \lVert C_i\mathbf{b}_i \rVert^2_2 + 2\sum_{i \neq j}^m \langle C_i\mathbf{b}_i, C_j\mathbf{b}_j \rangle
\end{align}

\noindent where the terms can also be precomputed and retrieved in $m$ table lookups. Thus, AQ has either a time ($2m$ vs. $m$ lookups) or memory overhead (for storing the quantized result of Eq.~\ref{eq:dnorm}) during distance computation with respect to PQ. Although this may sound as a major problem for AQ, it was shown in~\cite{aq} that sometimes the distortion error gain can be high enough that allocating memory from the code budget to store the result of Eq.~\ref{eq:dnorm} results in better recall \textit{and} faster distance computation compared to PQ/OPQ. This motivates us to look for better solutions to the AQ formulation.

\item For a given set of subcodebooks $C_i$ and a vector $\mathbf{x}$, encoding amounts to choosing the optimal set of codes $\mathbf{b}_i$ that minimize quantization error $\lVert \mathbf{x} - \sum_{i=1}^mC_i\mathbf{b}_i \rVert^2_2$. Unfortunately, without the orthogonality constraint the choice of $\mathbf{b}_i$ cannot be made independently in each subcodebook. This means that, in order to guarantee optimality, the search for the best encoding must be done over a combinatorial space of codewords. Moreover, it was shown in~\cite{aq} that this problem is equivalent to inference on a fully connected pairwise Markov Random Field, which is well-known to be NP-hard~\cite{mrfs-are-nphard}.

Since brute force search is not possible, one must settle for a heuristic search method. Beam search was proposed as a solution in~\cite{aq}, resulting in rather slow encoding. Beam search is done in $m$ iterations. At iteration $i$ the distance is computed from each of the $b$ candidate solutions to the set of $k \cdot (m-i)$ plausible candidates (in the $m-i$ codebooks that have not contributed to the candidate solution). At the end of the iteration we have $b^2$ candidate solutions, from which the top $b$ are kept as seeds for the next iteration~\cite{aq}. The complexity of this process is  $\mathcal{O}(m^2mbhd) = \mathcal{O}(m^3bhd)$, where $b$ is the search depth. As we will show, this makes the original solution of AQ impractical for very large databases.

\item Training consists of learning the subcodebooks $C_i$ and subcodebook assignments $\mathbf{b}_i$ that minimize expression~\ref{eq:compquant1}. A typical approach is to use coordinate descent by fixing the subcodebooks $C_i$ while updating the codes $\mathbf{b}_i$ (encoding), and later fixing $\mathbf{b}_i$ while updating $C_i$ (codebook update). As a side effect of slow encoding, we find that training is also very slow in AQ. While this might seem as a minor weakness of AQ (since training is usually done off-line, without tight time constraints), having faster training also means that for a fixed time budget we can handle larger amounts of training data. In the quantization setting, this means that we can use a larger sample to better capture the underlying distribution of the database.

In~\cite{aq}, codebook update is done by solving the over-constrained least-squares problem that arises from Eq.~\ref{eq:compquant} when holding $B$ fixed and solving for $C$. Fortunately, this decomposes into $d$ independent subproblems of $n$ equations over $mh$ variables~\cite{aq}. This corresponds to an optimal codebook update in the least squares sense.
We find that compared to encoding this step is rather fast, and thus focus on speeding up encoding.

\end{enumerate}






\section{Stacked Vector Quantizers}
\label{sect:haq}


\begin{figure*}[h]
\centering
\includegraphics[width=0.7\linewidth, trim=0 0 0 0, clip=true]{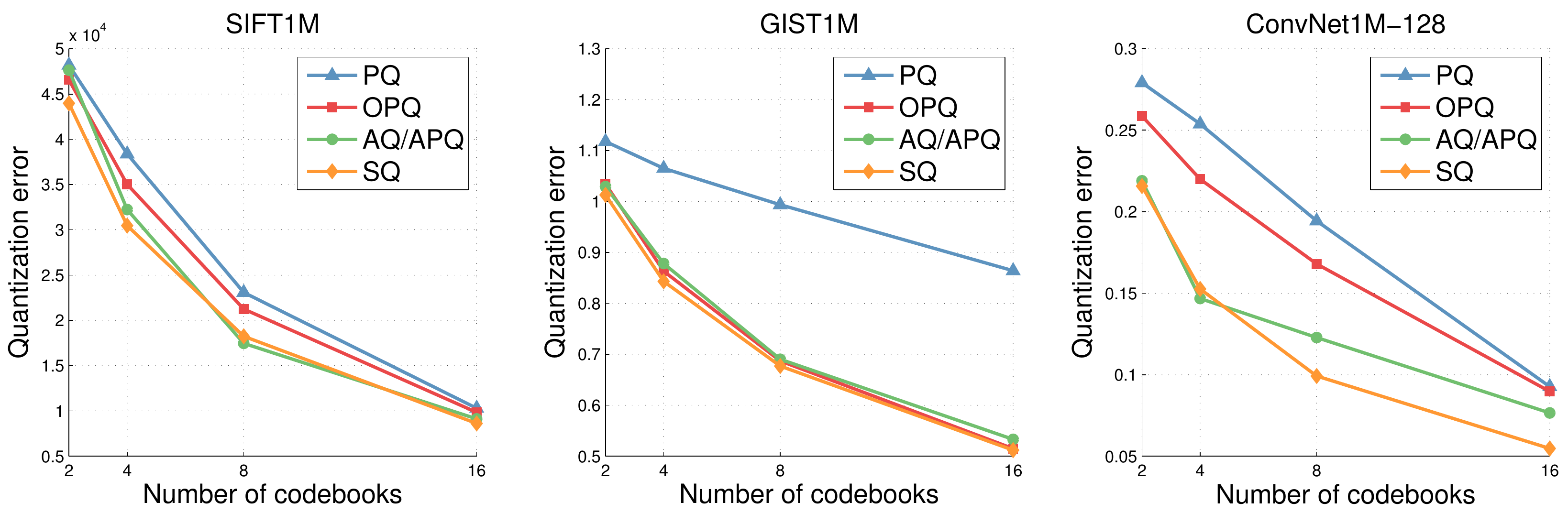}
\caption{Quantization error on SIFT1M, GIST1M and ConvNet1M 128. SQ shows the biggest performance advantage on deep features.}
\label{fig:distortions-1}
\end{figure*}

Within the subcodebook dependence-independence framework introduced in Section~\ref{sect:related}, we can see that PQ and OPQ assume subcodebook independence, while AQ embraces the dependence and tries to solve a more complex problem. As we will show next, there is a fertile middle ground between these approaches. 
We propose a \textit{hierarchical} assumption, which has the advantage of being fast to solve while maintaining the expressive power of AQ.
We now introduce our proposed approach to compositional quantization.

\vspace{-3mm}
\paragraph{Stacking Quantizers.} Due to the superior performance of AQ, we want to maintain its key property: subcodebook dependence. However, we look for a representation that can compete with PQ in terms of fast training and good scalability, for which fast encoding is essential. We propose to use a hierarchy of quantizers (see Figure~\ref{fig:overview}, left), where the vector is sequentially compressed in a coarse-to-fine manner.

\vspace{-3mm}
\paragraph{Encoding.} Fast encoding is at the heart of our approach. We assume that the subcodebooks $C_i$ have a hierarchical structure, where $C_1$ gives the coarsest quantization and $C_M$ the finest. Encoding is done greedily. In the first step, we choose the code $\mathbf{b}_1$ that most minimizes the quantization error $\lVert \mathbf{x} - C_1 \mathbf{b}_1\rVert^2_2$. Since all the subcodebooks are small, the search for $\mathbf{b}_1$ can be done exhaustively (as in k-means).

Next, we compute the first residual $\mathbf{r}_1 = \mathbf{x} - C_1\mathbf{b}_1$. We now quantize $\mathbf{r}_1$ using the codewords in $C_2$, choosing the one that minimizes the quantization error $\lVert \mathbf{r}_1 - C_2\mathbf{b}_2 \rVert^2_2$.
This process is repeated until we run out of codebooks to quantize residuals, with the last residual $\mathbf{r}_m$ being equal to the total quantization error (see Figure~\ref{fig:overview}, right). Now it is clear that we satisfy our first desired property, as the representation is additive in the encodings: $\mathbf{x} \approx \sum_{i=1}^m C_i \mathbf{b}_i$,  and the codewords all are $d$-dimensional (\ie, not independent of each other).

The complexity of this step is $\mathcal{O}(mhd)$ for $m$ subcodebooks, each having $h$ subcodewords, and a vector of dimensionality $d$. This corresponds to a slight increase in computation with respect to PQ ($\mathcal{O}(hd)$), but is much faster than AQ ($\mathcal{O}(m^3bhd)$). Given that encoding is only slightly more expensive than PQ, we can say that we have also achieved our second desired property.

\vspace{-3mm}
\paragraph{Initialization.} The goal of initialization is to create a coarse-to-fine set of codebooks. This can be achieved by simply performing k-means on $X$, obtaining residuals by subtracting the assigned codewords, and then performing k-means on the residuals until we run out of codebooks.

Formally, in the first step we obtain $C_1$ from the cluster centres computed by k-means on $X$, and we obtain residuals by subtracting $R_1 = X - C_1 B_1$. In the second step we obtain $C_2$ from k-means on $R_1$, and the residuals are refined to
$R_2 = R_1 - C_2 B_2$.
This process continues until we run out of codebooks (notice how this both is analogous to, and naturally gives rise to, the fast encoding proposed before). By the end of this initialization, we have an initial set of codebooks $C = [ C_1, C_2, \dots, C_m ]$ that have a hierarchical structure, and with which encoding can be performed in a greedy manner.

The computational cost of this step is that of running k-means on $n$ vectors $m$ times, \ie, $\mathcal{O}(mnhdi)$ for subcodebooks of size $h$, dimensionality $d$ and $i$ k-means iterations.
\vspace{-3mm}
\paragraph{Codebook refinement.} The initial set of codebooks can be further optimized with coordinate descent. This step is based on the observation that, during initialization, we assume that in order to learn codebook $C_i$ we only need to know codebooks $C_1, C_2, \dots, C_{i-1}$. However, after initialization all the codebooks are fixed. This allows us to fine-tune each codebook given the value of the rest.



Although it is tempting to use the least-squares-optimal codebook update proposed in~\cite{aq}, we have found that this tends to destroy the hierarchical subcodebook structure resulting from initialization. Without a hierarchical structure encoding cannot be done fast, which is one of the key properties that we wish to maintain. We therefore propose an ad hoc codebook refinement technique that preserves the hierarchical structure in the codebooks.

Let us define $\hat{X}$ as the approximation of $X$ from its encoding
\begin{equation}
\hat{X} = CB.
\end{equation}

Now, let us define $\hat{X}^{-i}$ as an approximation to the original dataset $X$ obtained using the learned codebooks $[ C_1, C_2, \dots, C_m ]$ and codes $B = [B_1^\top, B_2^\top, \dots, B_n^\top]^\top$ , \textit{except} for $C_i$, \ie,

\begin{equation}
\hat{X}^{-i} = \hat{X} - C_i B_i.
\end{equation}

We can now see that the optimal value of $C_i$ given the rest of the codebooksis obtained by running k-means on $X - X^{i-1}$, \ie, the residual after removing the contribution of the rest of the codebooks. Since we already know the cluster membership to $C_i$ (\ie, we know $B_i$) either from initialization or the previous iteration, we need to update only the cluster centres instead of restarting k-means (similar to how OPQ updates the codebooks given an updated rotation~\cite{opq, ckmeans}).

Enforcing codebook hierarchy is of the essence. Therefore, we run our codebook update in a top-down manner. We first update $C_1$ and update all codes. Next, we update $C_2$ and update codes again. We repeat the process until we have updated $C_m$, followed by a final update of the codes. Updating the codes after each codebook update ensures that the codebook hierarchy is maintained. A round of updates from codebooks 1 to $m$ amounts to one iteration of our codebook refinement.

The algorithm involves encoding using $m$ codebooks in the first pass, $m-1$ in the second pass, $m-2$ in the third pass and so on until only one set of codes is updated. This means that the time complexity of the codebook refinement procedure is quadratic in the number of codebooks. This is a significant increase with respect to PQ/OPQ, which are linear in $m$ during their training, but also represents an important reduction against the cubic scaling of AQ. Also, notice that the training usually has to be done only once with a small data sample, and database encoding remains efficient.

\begin{figure*}[h]
	\centering
	\includegraphics[width=0.95\linewidth, trim=0 0 0 0, clip=true]{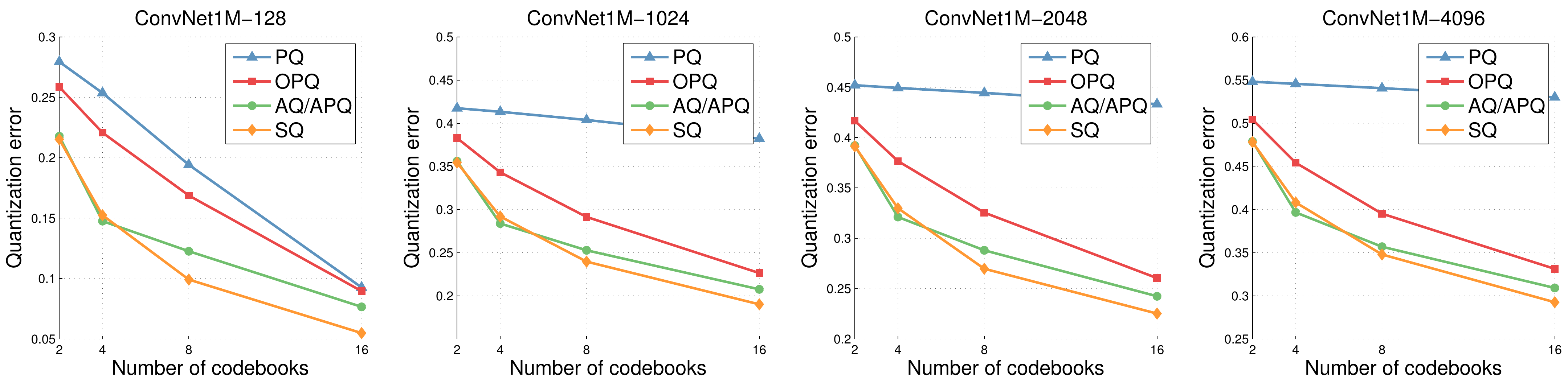}
	\caption{Quantization error on deep features with different dimensionalities. The non-independent approaches AQ and SQ clearly outperform PQ and OPQ. SQ achieves the best performance when using 8 and 16 codebooks (64 and 128 bits per feature) in all cases.}
	\label{fig:distortions-2}
\end{figure*}

\section{Experiments}
\label{sect:experiments}

Our main interest is to reduce quantization error because it has been demonstrated to lead to better retrieval recall, mean average precision and classification performance ~\cite{aq, opq, pq, ckmeans}. We also demonstrate two applications of our method: (i) approximate nearest neighbour search 
and (ii) classification performance with compressed features. In all our experiments we use codebooks of size $h=256$; this means that 2, 4, 8 and 16 codebooks generate codes of 16, 32, 64 and 128 bits.

\vspace{-3mm}
\paragraph{Datasets.} We test our method on three datasets. The first two are SIFT1M and GIST1M, introduced in~\cite{pq}. SIFT1M consists of 128-dimensional SIFT~\cite{sift} descriptors, and GIST1M consists of 960-dimensional GIST~\cite{gist} descriptors. Since hand-crafted features are consistently being replaced by features obtained from deep convolutional neural networks, we also consider a dataset of deep features: ConvNet1M-128. We obtained ConvNet1M-128 by computing 128-dimensional deep learning features on the ILSVRC-2012 training dataset~\cite{imagenet} using the CNN-M-128 network provided by Chatfield \etal~\cite{devil} and then subsampling equally at random from all classes. This network follows the architecture proposed by Zeiler and Fergus~\cite{visualizing}, with the exception that the last fully-connected layer was reduced from 4096 to 128 units. It has been shown that this intra-net compression has a minimal effect on classification performance~\cite{devil}, and exhibits state-of-the-art accuracy on image retrieval~\cite{gpuretrieval}. However, to the best of our knowledge we are the first to benchmark quantization techniques on deep learning features. We obtained the features from a central $224 \times 224$ image crop without further data augmentation. In the three datasets 100\,000 vectors are given for training, 10\,000 for query and 1\,000\,000 for database.

\vspace{-3mm}
\paragraph{Baselines.} We compare against 3 baselines. The first one is AQ as proposed by Babenko and Lempitsky~\cite{aq}, which consists of beam search for encoding and a least-squares codebook update in an iterative manner. As in~\cite{aq}, we set the beam search depth $b$ to 16 during training and to 64 for the database encoding. Although~\cite{aq} does not mention the number of iterations used during training, we found that 10 iterations reproduce the results reported by the authors and, as we will show, this is already several orders of magnitude slower than our approach. Since encoding scales cubically with the number of codebooks, for code lengths of 64 and 128 bits (8 and 16 codebooks respectively) we use the hybrid APQ algorithm suggested in~\cite{aq}, where the dataset is first preprocessed with OPQ, and then groups of 4 subcodebooks are refined independently with AQ. APQ was proposed for practical reasons, as otherwise AQ would require several days to complete given more than 4 subcodebooks: the need for this approximation starts to show the poor scalability of AQ. Since no code for AQ is available, we wrote our own implementation and incorporated the optimizations suggested in~\cite{aq}. We will make all our code available, including this baseline.

The second baseline is Optimized Product Quantization~\cite{opq, ckmeans}, which was briefly introduced in Section~\ref{sect:related}. We use the publicly available implementation by Norouzi \& Fleet\footnote{\url{https://github.com/norouzi/ckmeans}}, and set the number of optimization iterations to 100. The third baseline is Product Quantization~\cite{pq}. We slightly modified the OPQ code to create this baseline. We also use 100 iterations in PQ.

\begin{figure*}[h]
	\centering
	\includegraphics[width=0.9\linewidth, trim=0 0 0 0, clip=true]{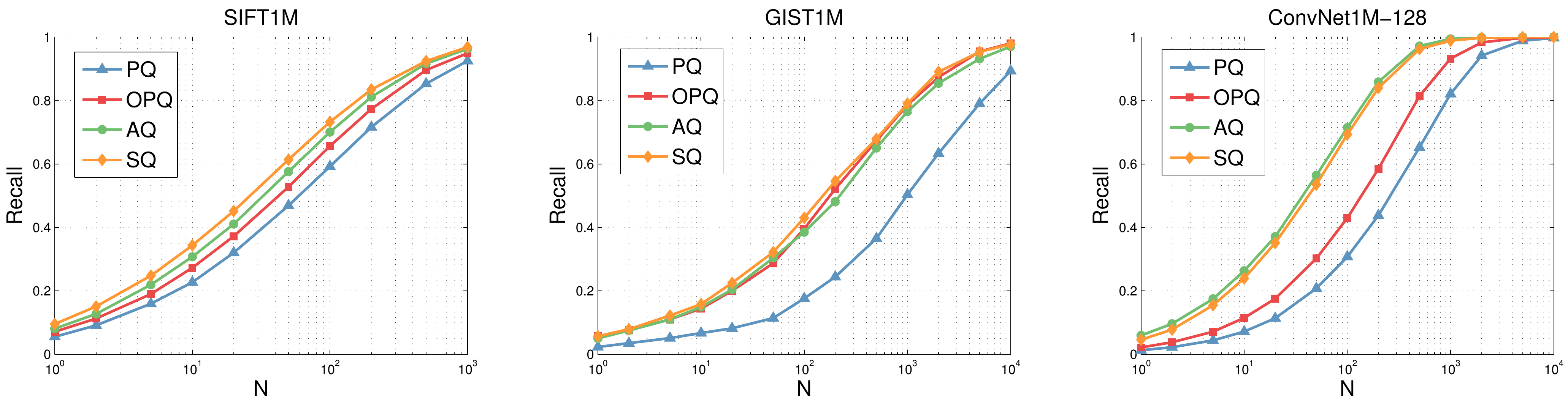}
	\caption{Recall@N on the SIFT1M, GIST1M and Convnet1M-128 datasets. Because of limited space, we consider only 32-bit codes (4 codebooks). We confirm previous observations~\cite{aq,ckmeans,opq} that correlate quantization error with nearest neighbour search performance: our method has the best recall for all values of $N$ in the SIFT1M and GIST1M datasets. SQ is slightly outperformed by AQ on ConvNet1M-128, but still performs much better than PQ and OPQ.  We show results for longer codes in the supplementary material.}
	\label{fig:recall}
\end{figure*}


\subsection{Quantization Error} Our main quantization results are shown on Figure~\ref{fig:distortions-1}. First, we observe that our method has a performance similar to AQ on SIFT1M and GIST1M. This is already good news, given the better scalability of our method. Moreover, we note that SQ obtains a large advantage on the deep features of ConvNet1M-128 when using 8 and 16 codebooks. We find this result rather encouraging, as deep features are likely to replace hand-crafted descriptors such as SIFT and GIST in the foreseeable future.

OPQ achieves a large gain compared to PQ in GIST1M, and this gap is only slightly improved by AQ and SQ. Since both SIFT1M and ConvNet1M-128 have low dimensionality (128), and GIST1M has high-dimensional descriptors (960), it remains unclear whether the advantages of AQ and SQ are only restricted to low-dimensional descriptors. We investigate this question by benchmarking the methods on 1024-, 2048- and 4096-dimensional deep features obtained in a similar manner to ConvNet1M-128, but using using the CNN-M-1024, CNN-M-2048 and CNN-M networks from~\cite{devil} respectively. The quantization results on these datasets are shown on Figure~\ref{fig:distortions-2}. While the PQ-to-OPQ gap is still present for high-dimensional features, we see that AQ and SQ maintain a performance gap from OPQ similar to that observed on the 128-dimensional features. Moreover, our method remains the clear winner for 8 and 16 codebooks, and largely competitive with AQ for 4 codebooks. These results suggest that codebook independence hurts the compression of deep features particularly badly and motivates more research of compositional quantization methods that follow the formulation of expression~\ref{eq:compquant1}.

\subsection{Approximate Nearest Neighbour Search} We demonstrate the performance of our method on fast search of $K$ nearest neighbours with recall@N curves~\cite{pq}. These curves represent the probability of the true $K$ nearest neighbours being in a retrieved list of $N$ neighbours for varying $N$. We set $K=1$ and observe little variability for other values. Our main results are shown on Figure~\ref{fig:recall}.
As expected, lower quantization error lets us achieve higher recall on SIFT1M and GIST1M, although on GIST1M OPQ and AQ achieve very competitive performance. On ConvNet1M-128, our method was slightly outperformed by AQ; however, this trend is reversed for longer codes, consistent with the quantization error of Fig.~\ref{fig:distortions-1}. We show results on longer codes in the supplementary material.

\begin{figure}
\centering
\includegraphics[width=1\linewidth]{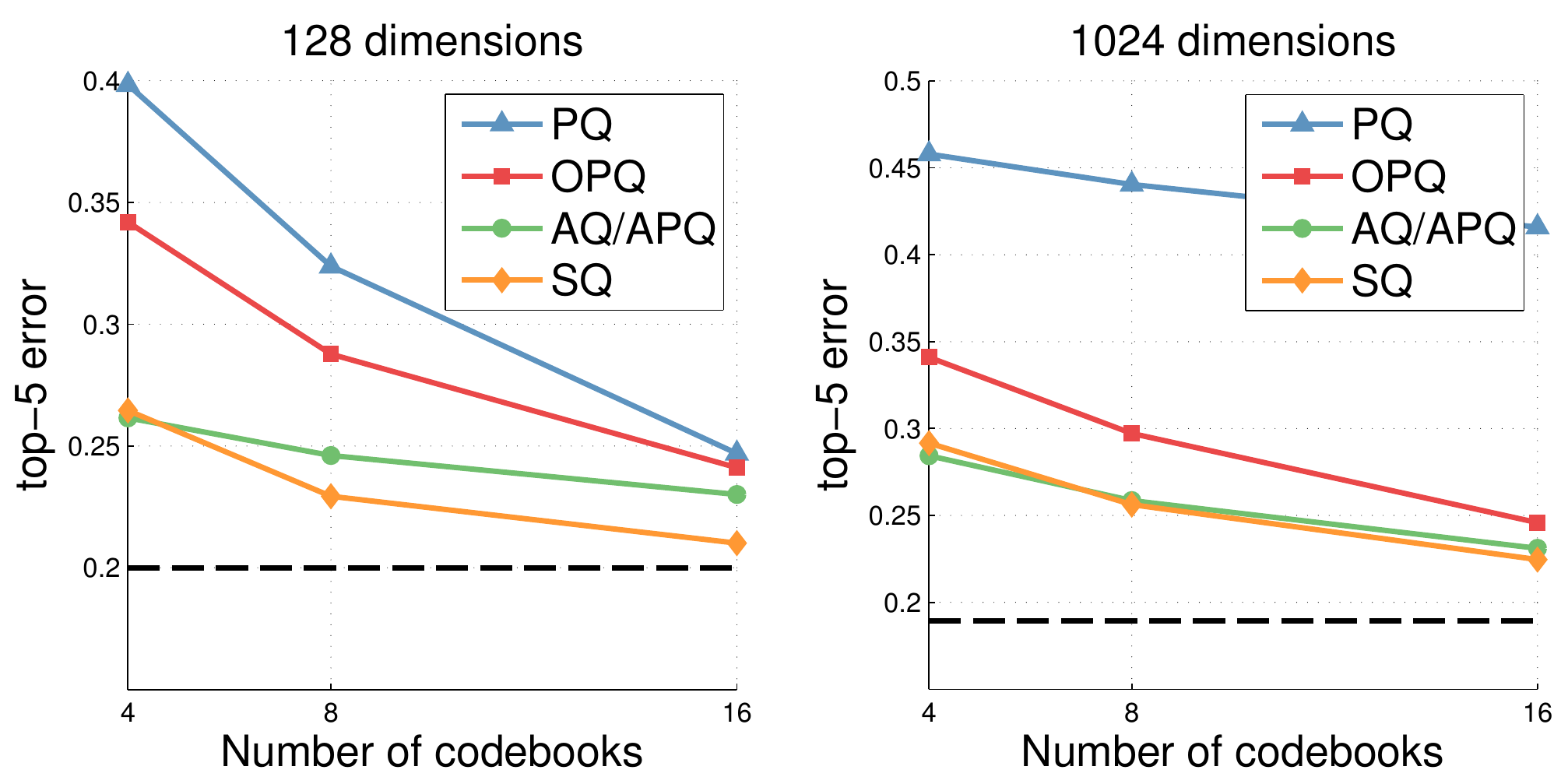}
\caption{Top-5 classification error on the ILSVRC-2012 dataset as a function of compression. The dotted black line corresponds to performance without compression. The left pane shows performance using 128-dimensional deep features, and the right pane shows performance for 1024-dimensional deep features.}
\label{fig:classification}
\end{figure}

\subsection{Large-Scale Object Classification} We study the trade-off in classification performance vs. compression rate on the ILSVRC-2012 dataset using deep learning features. We trained a linear SVM on the 1.2 million uncompressed examples provided, and preprocessed the features with L2 normalization, which was found to improve performance in~\cite{devil}. The 50\,000 images in the validation set were preprocessed similarly and compressed before evaluation. This scenario is particularly useful when one wants to search for objects in large unlabelled datasets~\cite{queries, picodes}, and in retrieval scenarios where classifiers are applied to large collections of images in search for high scores~\cite{gpuretrieval, high}. Notice that in this scenario, the only operation needed between the support vectors and the database descriptors is a dot product; as opposed to distance computation, this can be done with $m$ lookups in AQ and SQ, the same as for PQ and OPQ. We report the classification error taking into account the top 5 predictions.

Classification results are shown on Figure~\ref{fig:classification}. We observe a similar trend to that seen in our quantization results, with PQ and OPQ consistently outperformed by AQ and SQ. Using 128-dimensional features our method performs similarly to AQ using 4 codebooks, but shows better performance for larger code sizes. Using 1024-dimensional features AQ and SQ are practically equivalent but, curiously, it seems like the 128-dimensional features are more amenable to compression: for all compression rates the 128-dimensional features outperform the 1024-dimensional features ($[0.2646, 0.2293, 0.2101]$ vs. $[0.2917, 0.2562, 0.2246]$ in top-5 error), even though when uncompressed the 1024-dimensional features perform slightly better ($0.1999$ vs. $0.1893$). This suggests that, if quantization is planned as part of a large-scale classification pipeline, low-dimensional features should be preferred over high-dimensional ones. It is also noticeable that for extreme compression rates (\eg, 32 bits) PQ and OPQ have error rates in the 35-45\% range, while AQ and SQ degrade more gracefully and maintain a 25-30\% error rate.

\subsection{Running times}

\begin{figure}
\centering
\includegraphics[width=1\linewidth]{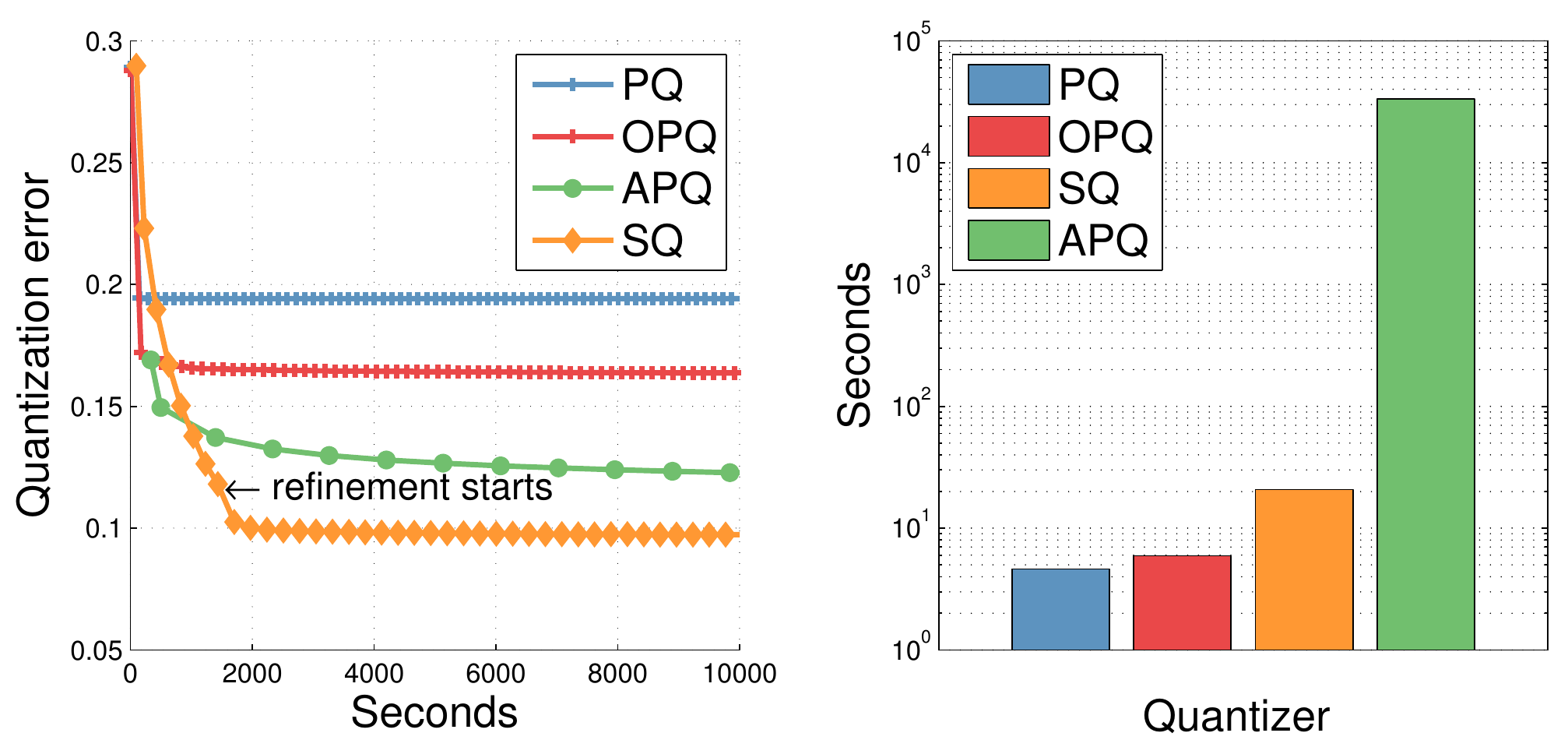}
\caption{Left: Training time vs. Quantization error of the benchmarked methods on the ConvNet1M-128 training dataset (100K features). For clarity, we plot each 50 iterations for PQ and OPQ and each 25 iterations for SQ after initialization. PQ and OPQ complete 100 iterations after 286 and 336 seconds respectively (4.8 and 5.6 minutes), SQ takes ${\sim} 1500$ seconds for initialization and ${\sim} 1000$ seconds for 100 iterations of codebook refinement (42 minutes in total). APQ takes ${\sim} 2.7$ hours for 10 iterations. Right: Encoding of the database set of 1M features. PQ and OPQ take ${\sim} 5$ seconds, SQ ${\sim} 20$ seconds, and APQ ${\sim} 9.2$ hours.}
\label{fig:time}
\end{figure}

Figure~\ref{fig:time} shows the running time for training and database encoding for PQ/OPQ, APQ and SQ on the ConvNet1M-128 dataset using 8 codebooks (64 bits). All measurements were taken on a machine with a 3.20 GHz processor using a single core. We can see that SQ obtains most of its performance advantage out of initialization, but codebook refinement is still responsible for a 20\% decrease to the final quantization error (0.12 to 0.10). We also see that APQ largely improves upon its OPQ initialization, but these iterations are extremely expensive compared to PQ/OPQ, and 3 iterations take almost as much computation as the entire SQ optimization. Beyond training (which arguably is not too big of a problem, since it only has to be done once), encoding the database with the learned codebooks is extremely expensive with APQ (9.2 hours), while for PQ/OPQ and SQ it stays in the 5-20 second range. Projecting these numbers to the encoding of a dataset with 1 billion features such as SIFT1B~\cite{pq} suggests that PQ/OPQ would need about 1.5 hours to complete, and SQ would need around 6 hours; however, APQ would need around 1.05 years (!). Although all these methods are highly parallelizable, these numbers highlight the importance of fast encoding for good scalability.



\section{Conclusions and future work}
\label{sect:future}
We have introduced Stacked Quantizers as an effective and efficient approach to compositional vector compression. After analyzing PQ and AQ in terms of their codebook assumptions, we derived a method that combines the best of both worlds, being only slightly more complex than PQ, while maintaining the representational power of AQ. We have demonstrated state-of-the-art performance on datasets of SIFT, GIST and, perhaps most importantly, deep convolutional features.

Future work will look at the integration of our pipeline with non-exhaustive indexing techniques such as the inverted file~\cite{pq} or the inverted multi-index~\cite{inverted}. We also plan to investigate the use of optimization approaches that have proven useful in network-like architectures such as stochastic gradient descent and conjugate gradient.

\vspace{-3mm}
\paragraph{Acknowledgements. } 
This work was supported in part by the Natural Sciences and Engineering Research Council of Canada (NSERC) and the Institute for Computing, Information and Cognitive Systems (ICICS) at UBC, and enabled in part by WestGrid and Compute / Calcul Canada.

{\footnotesize
\bibliographystyle{ieee}
\bibliography{egbib}
}

\end{document}